\documentclass[preprint,12pt]{elsarticle}

\usepackage{amssymb}
\usepackage{amsmath}
\usepackage{lineno}
\usepackage{subfigure}
\usepackage{kotex}
\usepackage{algorithmic}
\usepackage[normalem]{ulem}
\useunder{\uline}{\ul}{}
\usepackage{makecell}
\usepackage[linesnumbered,ruled,vlined]{algorithm2e}
\usepackage{color}
\usepackage[hang,flushmargin]{footmisc} 
\usepackage{graphicx}
\usepackage{booktabs}
\usepackage{multirow}
\usepackage{pifont}
\usepackage{hyperref}

\AtBeginDocument{%
  \providecommand\BibTeX{{%
    \normalfont B\kern-0.5em{\scshape i\kern-0.25em b}\kern-0.8em\TeX}}}

\makeatletter
\def\ps@pprintTitle{%
  \let\@oddhead\@empty
  \let\@evenhead\@empty
  \let\@oddfoot\@empty
  \let\@evenfoot\@empty
}
\makeatother
\begin{document}
\begin{frontmatter}


\title{Detecting the Undetectable: Enhancing Unsupervised time series Anomaly Detection via Active Learning}
\author[1]{Seung Hun Han}
\ead{seunghun.han@lgcns.com}

\author[2]{Hyeongwon Kang}
\ead{hyeongwon\_kang@korea.ac.kr}

\author[3]{Jinwoo Park}
\ead{jinwoo\_park@snu.ac.kr}

\author[3]{Pilsung Kang\corref{cor1}}
\ead{pilsung_kang@snu.ac.kr}

\address[1]{LG CNS, 71 Magokjungang 8-ro, Gangseo-gu, Seoul, Republic of Korea}

\address[2]{Department of Industrial \& Management Engineering, Korea University, 126-16 Anam-dong 5-ga, Seongbuk-gu, Seoul, Republic of Korea}

\address[3]{Department of Industrial Engineering, Seoul National University, Gwanak-ro 1, Gwanak-gu, Seoul, Republic of Korea}

\cortext[cor1]{Corresponding author}

\begin{abstract}
Despite the increasing sophistication of industrial AI systems, the ability to reliably detect subtle and noisy anomalies in complex time series data remains a critical yet unresolved challenge. 
In large-scale industrial applications, labeling time series data is often prohibitively expensive and time-consuming, making unsupervised learning a practical and widely adopted approach. 
However, existing unsupervised methods frequently struggle to distinguish near-normal anomalies from normal patterns and are vulnerable to noise contamination within normal samples. 
To address these limitations, we propose a novel framework that leverages active learning to iteratively enhance the performance of unsupervised models. 
Our framework's core contributions are (1) a masked time-series reconstruction feedback strategy that forces the model to learn robust temporal dependencies, and (2) a minimax learning strategy that promotes robustness by differentially treating normal and abnormal samples. 
This process encourages the model to better capture the dynamics of subtle and noisy patterns. The proposed framework is evaluated across 28 test cases involving four multivariate time-series datasets and seven unsupervised backbone models. 
Experimental results demonstrate a 12.39\% improvement in AUC compared to the original models, confirming that our method can be readily integrated into existing unsupervised reconstruction-based anomaly detection systems to significantly enhance their performance.
\end{abstract}

\end{frontmatter}



\section{Introduction}
\label{sec:Introduction}

In today's increasingly automated and sensor-rich industrial environments, vast streams of time series data are continuously generated, capturing the dynamic behaviors of machines, processes, and entire systems. Amid this data deluge, even minor anomalies can signal critical issues such as equipment failure, cyberattacks, or financial risk, underscoring the importance of timely and accurate detection. time series anomaly detection, which aims to identify patterns that deviate significantly from normal behavior, has therefore become a cornerstone of operational reliability and industrial risk mitigation \cite{RN24}. As modern systems grow more complex and interconnected, the need for intelligent and scalable anomaly detection tools has become more urgent than ever \cite{RN24, LSTM_VAE}. This growing demand has spurred intense research activity focused on developing effective methods for monitoring and detecting anomalies in multivariate time series data \cite{RN24, VAE_GAN, LIAD2021}.

Recent research on time series anomaly detection has increasingly emphasized unsupervised learning methods, primarily due to the practical challenges of obtaining high-quality labeled data in real-world applications \cite{LSTM_VAE, USAD, OmniAnomaly}. Annotating time series data is not only labor-intensive and time-consuming, but also requires substantial domain expertise and a nuanced understanding of temporal dependencies and inter-variable relationships \cite{Arslan23, zhaoetal, choiad}. As a result, most studies have focused on modeling normal behavioral patterns without labeled anomalies, identifying deviations as potential outliers \cite{DCdetector, VTT}. Traditional unsupervised techniques, including Local Outlier Factor (LOF) \cite{LOF}, One-Class SVM \cite{OCSVM}, Support Vector Data Description (SVDD) \cite{SVDD}, and Isolation Forest \cite{IF}, offer lightweight solutions but struggle with high-dimensional or complex time series data due to their limited capacity to capture temporal structure \cite{LiAD, SuAD, Thooc}. To address these limitations, deep learning-based models—such as Convolutional Neural Networks (CNNs) \cite{CNN}, Recurrent Neural Networks (RNNs) \cite{RNN}, and Transformers \cite{Transformer}—have gained traction for their ability to learn complex sequential dynamics. This shift has spurred widespread interest in applying deep learning to unsupervised time series anomaly detection, particularly in large-scale industrial settings \cite{USAD, Thooc}.

Despite the progress enabled by unsupervised deep learning models, fundamental limitations remain. One key challenge lies in their susceptibility to noise: normal data with minor fluctuations is frequently misclassified as anomalous, resulting in a high false positive rate \cite{sensitivehue}. Conversely, subtle anomalies that closely mimic normal patterns often go undetected, giving rise to false negatives \cite{Imdiffusion}. At the core of these limitations is the difficulty unsupervised models face in distinguishing between noisy normal samples and true anomalies—particularly in high-dimensional time series data with complex temporal dynamics and inter-variable dependencies \cite{PAK, TSADFlawed, MSCRED}. In response, recent research has sought to improve the fidelity of normal pattern learning and to enhance robustness against such confounding factors, aiming to bridge the performance gap in practical deployments \cite{USAD, VTT, anomaly_transformer}. This limitation is not trivial; it is fundamental to the unsupervised paradigm, which lacks semantic grounding. This ambiguity inevitably leads to a high false positive rate or critical false negatives, creating a performance ceiling that pure unsupervised methods cannot overcome \cite{Active_MTSAD}. This necessitates a hybrid approach that strategically incorporates minimal human expertise.

To overcome the limitations of purely unsupervised models, recent research has explored hybrid approaches that incorporate limited supervision to guide anomaly detection. In particular, active learning has emerged as a promising strategy for selectively annotating the most informative samples from unlabeled time series data, thereby enhancing model performance with minimal labeling effort \cite{Active_MTSAD, AMAD, TS_ACtive2, LittleHelps}. By focusing annotation efforts on samples that are most uncertain or representative, active learning directly addresses the challenges unsupervised models face, especially in distinguishing noisy normal patterns from subtle anomalies \cite{activelearning}. This paradigm offers a practical compromise between fully supervised and unsupervised learning, and is particularly well-suited for time series anomaly detection, where exhaustive labeling is rarely feasible and unsupervised methods alone may fall short of desired accuracy \cite{Active_MTSAD}.

While active learning has shown promise in enhancing time series anomaly detection, its integration into deep learning-based unsupervised reconstruction models remains relatively underexplored. Existing approaches often employ traditional machine learning models as backbones \cite{LittleHelps}, or rely on supervised pre-training before applying active learning strategies \cite{AMAD, TS_ACtive2}. Among them, Active-MTSAD \cite{Active_MTSAD} most closely aligns with our objective by combining deep unsupervised models with selective supervision through active learning. It adopts a pseudo-labeling strategy in which selected queries from the unlabeled pool are annotated and used to train the model via metric learning. However, this reliance on pseudo-labels can be problematic, particularly when the limited labeled samples fail to capture the diversity of real-world anomaly patterns. Furthermore, Active-MTSAD assumes that each query round includes at least one true anomaly, a condition that is often unmet in industrial settings where anomalies are both rare and stochastic. These limitations reduce the method’s robustness and hinder its practical applicability in large-scale industrial scenarios.

We propose an active learning-based framework designed to enhance the performance of unsupervised reconstruction models for time series anomaly detection. Our core novelty lies not in the selection of queries, but in how the model learns from them. Our approach focuses on enabling the model to better distinguish between noisy normal samples and subtle anomalies. Specifically, we introduce two key components: (1) a masked time series reconstruction feedback strategy that strengthens the model's understanding of temporal dependencies, and (2) a minimax loss that treats normal and anomalous samples separately to guide the learning of discriminative temporal patterns. To source informative samples for this sophisticated feedback mechanism, we employ an effective hybrid query strategy, adapting principles from prior work \cite{Active_MTSAD} (e.g., top-k and interval sampling) to capture both noisy normal data with high scores and near-normal anomalies with lower scores. Unlike prior methods that rely on pseudo-labeling and structural modifications, our framework is directly applicable to a wide range of reconstruction-based unsupervised models. Our main contributions are as follows:

\begin{itemize}
\item \textbf{Extraction of Noisy and Anomalous Patterns}: We design a feedback mechanism that extracts noisy normal and near-normal anomalous samples from unlabeled data using an active learning strategy. These queries are labeled by an oracle and used to further train the backbone model. A minimax loss function—applied separately to normal and anomalous queries—enhances detection accuracy.
\item \textbf{Temporal Reconstruction via Pretext Task}: To capture complex temporal dynamics, we introduce a masked reconstruction task within the feedback strategy. By reconstructing randomly masked segments of the input, the model is encouraged to learn richer temporal representations that improve anomaly discrimination.
\item \textbf{Model-Agnostic Framework}: Our framework is architecture-agnostic and can be seamlessly integrated into existing deep unsupervised reconstruction models without structural modifications. It consistently improves performance across diverse backbones.
\end{itemize}

The remainder of this paper is structured as follows. Section~\ref{sec:RelatedWorks} reviews related work on unsupervised time series anomaly detection and the application of active learning in time series analysis. Section~\ref{sec:Proposed Method} presents the proposed methodology, detailing each component of the framework. Section~\ref{sec:Experimental Settings} describes the experimental setup and evaluation protocol. Section~\ref{sec:Experimental Results} reports the results of extensive experiments and ablation studies. Section~\ref{sec:Analysis} provides a comparative analysis of anomaly detection performance with and without active learning. Finally, Section~\ref{sec:Conclusion} concludes the paper with a summary of key findings and future research directions.

\section{Related Works}
\label{sec:RelatedWorks}

\subsection{Unsupervised Multivariate time series Anomaly Detection}
\label{sec:Unsupervised Multivariate time series Anomaly Detection}
Unsupervised time series anomaly detection methods can be broadly categorized into four groups: density estimation-based, clustering-based, prediction-based, and reconstruction-based approaches. Each category offers a distinct strategy for modeling normal behavior and identifying deviations as anomalies.

\textbf{Density Estimation-Based Approaches.}
These methods identify anomalies as data points located in low-density regions of the data distribution. Local Outlier Factor (LOF) \cite{LOF}, for instance, classifies a point as anomalous if its local density significantly deviates from its neighbors. DAGMM \cite{DAGMM} extends this idea by integrating Gaussian Mixture Models (GMMs) with an autoencoder, jointly optimizing for reconstruction loss and density estimation.

\textbf{Clustering-Based Approaches.}
Clustering-based methods assume that normal samples form dense clusters, while anomalies reside far from cluster centroids. Deep-SVDD \cite{DeepSVDD}, a deep learning extension of Support Vector Data Description (SVDD) \cite{SVDD}, maps data into a latent space and encloses normal samples within a compact hypersphere. Samples falling outside this boundary are considered anomalies.

\textbf{Prediction-Based Approaches.}
Prediction-based techniques model normal temporal patterns and identify anomalies as instances with large prediction errors. These methods vary by the prediction backbone used. Traditional statistical models like ARIMA \cite{ARIMA} capture linear trends, while deep learning models such as LSTM-AD \cite{LSTM-AD} and CNN-based DeepANT \cite{DeepANT} are designed to learn nonlinear temporal dependencies from multivariate sequences.

\textbf{Reconstruction-Based Approaches.}
These models aim to learn a compressed representation of normal sequences, reconstructing them accurately while failing to reconstruct anomalies. LSTM-VAE \cite{LSTM_VAE} leverages a Variational Autoencoder to model the distribution of normal data and reconstruct samples from latent representations. USAD \cite{USAD} employs a dual-autoencoder structure to minimize reconstruction error, while OmniAnomaly \cite{OmniAnomaly} combines a Stacked Recurrent Autoencoder (SRAE) with a VAE to jointly model temporal dynamics and probabilistic structure.

Recent methods adopt Transformer architectures to better capture long-range dependencies in time series data. Anomaly Transformer \cite{anomaly_transformer} introduces a mechanism based on association discrepancy, detecting anomalies by evaluating the inconsistency of attention patterns. This reflects the observation that anomalous samples often fail to form strong associations with normal timestamps due to their rarity. Variable Temporal Transformer (VTT) \cite{VTT} further enhances temporal modeling by replacing standard self-attention with a temporal self-attention mechanism, allowing it to better capture variable dependencies across time.

\subsection{Active Learning in time series Anomaly Detection}
\label{sec:Active Learning in time series Anomaly Detection}
Recent efforts have explored the application of active learning to enhance time series anomaly detection, particularly in scenarios where labeling is costly and anomalies are rare. SLA-VAE \cite{TS_ACtive2} integrates active learning with LSTM-VAE by minimizing the Evidence Lower Bound (ELBO) for normal data while maximizing it for anomalies, enabling uncertainty estimation for query selection. AMAD \cite{AMAD} adopts a conventional active learning pipeline, beginning with a supervised training phase and iteratively refining the model by querying high-uncertainty samples. In another approach, Little Helps \cite{LittleHelps} uses an unsupervised Isolation Forest to generate anomaly scores and perform weighted feedback based on tree performance, operating in a feedback-driven unsupervised learning setting.
RLVAL \cite{reinforce_AL} proposes a novel framework for univariate time series that combines a Deep Reinforcement Learning (DRL) agent with a Variational Autoencoder (VAE) and Active Learning, utilizing an LSTM network to effectively model sequential dependencies. The core objective of this method is to overcome the limitations of static models, particularly in adapting to new anomaly types. By leveraging DRL and Active Learning, the framework aims to detect novel anomaly classes using minimal labeled data, addressing the common challenges of manual parameter tuning. While these approaches have advanced the integration of active learning in anomaly detection, they typically rely on traditional machine learning backbones or assume a semi-supervised initial setup, conditions that differ from the fully unsupervised reconstruction-based models considered in this study.

Among prior works, Active-MTSAD \cite{Active_MTSAD} is the most closely aligned with our research, as it also combines deep unsupervised models with active learning for time series anomaly detection. Both approaches acknowledge the limitations of pure unsupervised methods and employ a hybrid query strategy, which combines high-score sampling and diverse interval sampling to select informative samples.
However, our framework's novelty lies not in the sampling strategy, but in the fundamental design of the feedback mechanism, learning objective, and label handling, which directly address key limitations of Active-MTSAD. First, Active-MTSAD employs a pseudo-labeling strategy, where labels from a small query set are extrapolated to the entire unlabeled pool to guide metric learning. This reliance on pseudo-labels is problematic, as inaccuracies from the initial, small set of queries can propagate and amplify during training, potentially misleading the model. To address this, our framework avoids pseudo-labeling altogether. We exclusively use the high-fidelity, oracle-verified labels from the query set for supervised feedback, ensuring that the model learns from reliable ground-truth information.
Second, Active-MTSAD uses metric learning to pull similar samples together and push dissimilar ones apart in a latent space. In contrast, we introduce a masked time series reconstruction feedback strategy. This pretext task forces the backbone model to learn the underlying temporal dynamics and complex dependencies within the challenging `noisy-normal' and `subtle-anomaly' sequences themselves, rather than just learning their relative distances in a latent space.
Third, Active-MTSAD's feedback relies on an assumption that each query round includes at least one true anomaly \cite{Active_MTSAD}, a condition often unmet in real-world industrial settings with sparse anomalies. Our minimax objective is inherently robust to this scenario. If a query batch contains zero anomalies, the ``maximize" phase is simply skipped, and the model robustly refines its understanding of `noisy normal' samples, avoiding the instability or failure modes of the prior method.

In summary, while we adapt a similar query selection component, our framework introduces a fundamentally different feedback architecture that is more robust, avoids error propagation, and is better suited for practical, large-scale industrial scenarios.

\section{Proposed Method}
\label{sec:Proposed Method}

\subsection{Limitations of Unsupervised Anomaly Detection Models}
\label{sec:Motivation}

\begin{figure*}[!t]
    \centering  
    \includegraphics[width=13cm]{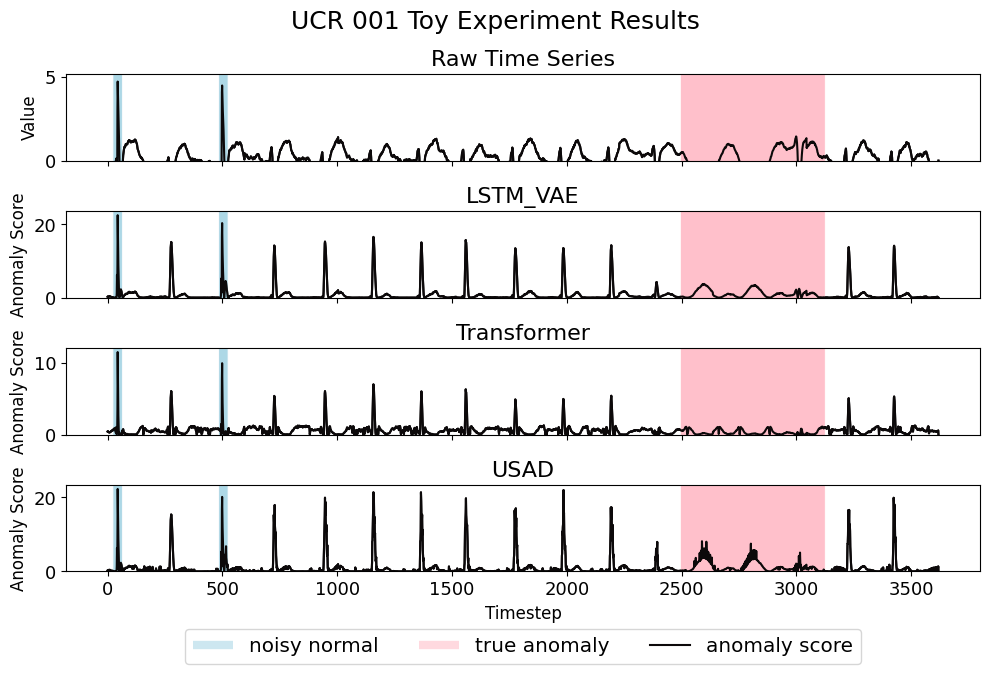}
    \caption{Raw time series of UCR 001 and anomaly score of each model where pink segments are anomalous timesteps}
    \label{fig:UCR_TOY}
\end{figure*}

Despite the advances in unsupervised learning, existing reconstruction-based anomaly detection models often struggle to reliably distinguish between noisy normal samples and subtle anomalies that closely resemble normal patterns. To empirically illustrate this issue, we evaluated three representative models, LSTM-VAE \cite{LSTM_VAE}, Transformer \cite{Transformer}, and USAD \cite{USAD}, using a benchmark univariate time series dataset from the UCR archive \cite{UCR}, which includes both noisy normal and near-normal anomalous segments.

As shown in \autoref{fig:UCR_TOY}, the blue segment denotes normal data, while the pink segment indicates anomalous data. Notably, all three models assign high anomaly scores to the blue segment due to abrupt but benign fluctuations, potentially leading to false positives. Conversely, the red segment, although anomalous, exhibits only subtle deviations, resulting in low anomaly scores and thus false negatives.

This example highlights a critical limitation of unsupervised approaches: their reliance on reconstruction or prediction errors as indirect proxies for anomaly likelihood makes them sensitive to noise and less effective in capturing fine-grained deviations. Prior studies \cite{choiad, Thooc, MSCRED} have reported similar findings, showing that these models often fail to generalize in complex, real-world time series settings.

To address this challenge, active learning has gained traction as a promising strategy. By selectively labeling noisy normal and near-normal anomalous samples, active learning supplements the unsupervised training process with minimal supervision, guiding the model to better discriminate between similar-looking patterns. This targeted feedback helps correct common misclassifications and improves model robustness with a modest labeling budget.

Motivated by these observations, we propose an active learning-based framework that explicitly incorporates such challenging samples into the training loop. Our method aims to bridge the gap between unsupervised generality and supervised precision, thereby enhancing anomaly detection performance in real-world scenarios where noise and ambiguity are prevalent.

\begin{figure*}[!t]
    \centering  
    \includegraphics[width=14cm]{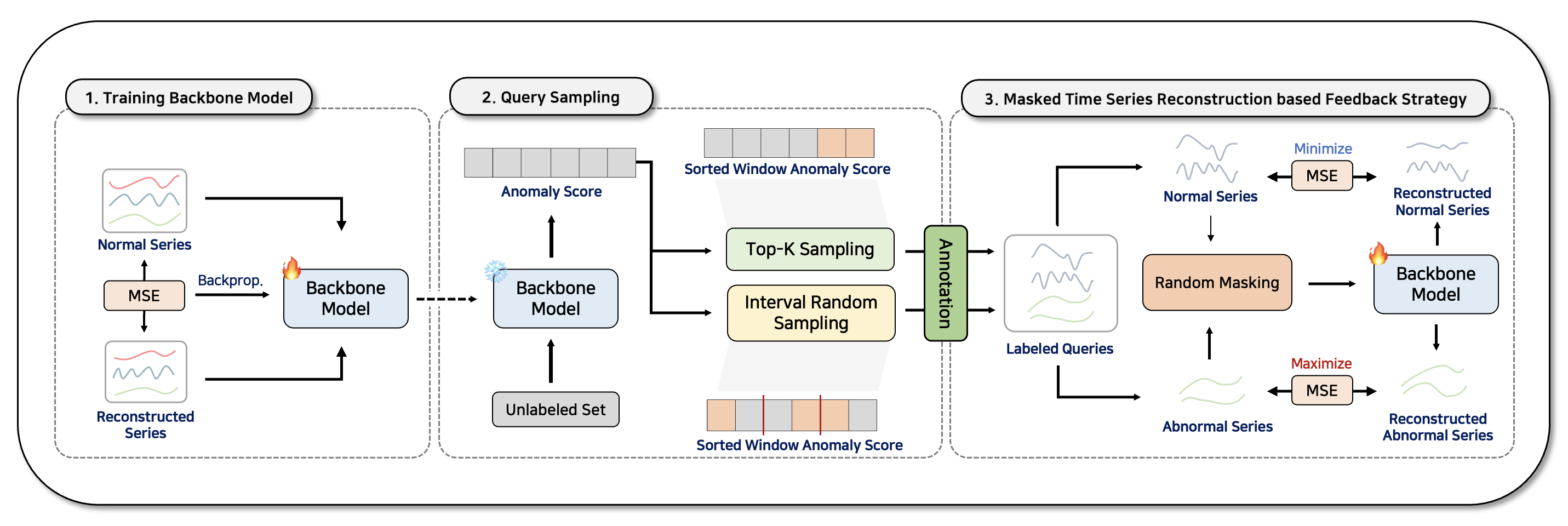}
    \caption{Overall architecture of proposed framework}
    \label{fig:Framework}
\end{figure*}

\subsection{Overall Framework}
\label{sec:Overall Framework}

In unsupervised multivariate time series anomaly detection, the input data is denoted by $\mathbb{X}_{\text{Total}} \in \mathbb{R}^{S \times C}$, where $S$ is the total sequence length and $C$ represents the number of channels. A sliding window approach is employed to generate fixed-length subsequences. At each timestep $t$, a sequence of length $L$ is formed as $\mathbb{X}_{t} = {X_{t-L+1}, X_{t-L+2}, \ldots, X_t}$, with each $X_l \in \mathbb{R}^{1 \times C}$. These sequences are input to the model, which is trained to reconstruct them as $\hat{\mathbb{X}}_{t} = {\hat{X}_{t-L+1}, \hat{X}_{t-L+2}, \ldots, \hat{X}_t}$, where each $\hat{X}_l \in \mathbb{R}^{1 \times C}$. The model learns to minimize the reconstruction loss between the original and reconstructed sequences.

During inference, the anomaly score $S_t$, based on Mean Squared Error (MSE), for each timestep $t$ is computed as the average reconstruction error over the sequence. Points with anomaly scores exceeding a predefined threshold are classified as anomalous. The reconstruction loss is defined as follows:
\begin{align} 
    \label{eq:MSEloss} 
    \text{loss} (\mathbb{X}_{t}, \hat{\mathbb{X}}_{t}) = \frac{1}{L} \sum_{l=0}^{L-1} \left(X_{t-l} - \hat{X}_{t-l}\right)^2 
\end{align}
The threshold is selected based on the Best-F1 criterion to maximize detection performance on the evaluation dataset.

As illustrated in \autoref{fig:Framework}, the proposed framework operates in three stages:

\textbf{Stage 1: Unsupervised Pretraining.}
The backbone model is first trained in an unsupervised setting using normal time series data. Given input sequences $\mathbb{X} \in \mathbb{R}^{L \times C}$, the model learns to reconstruct them by minimizing reconstruction error, effectively capturing normal temporal patterns.

\begin{figure*}[!t]
    \centering  
    \includegraphics[width=10cm]{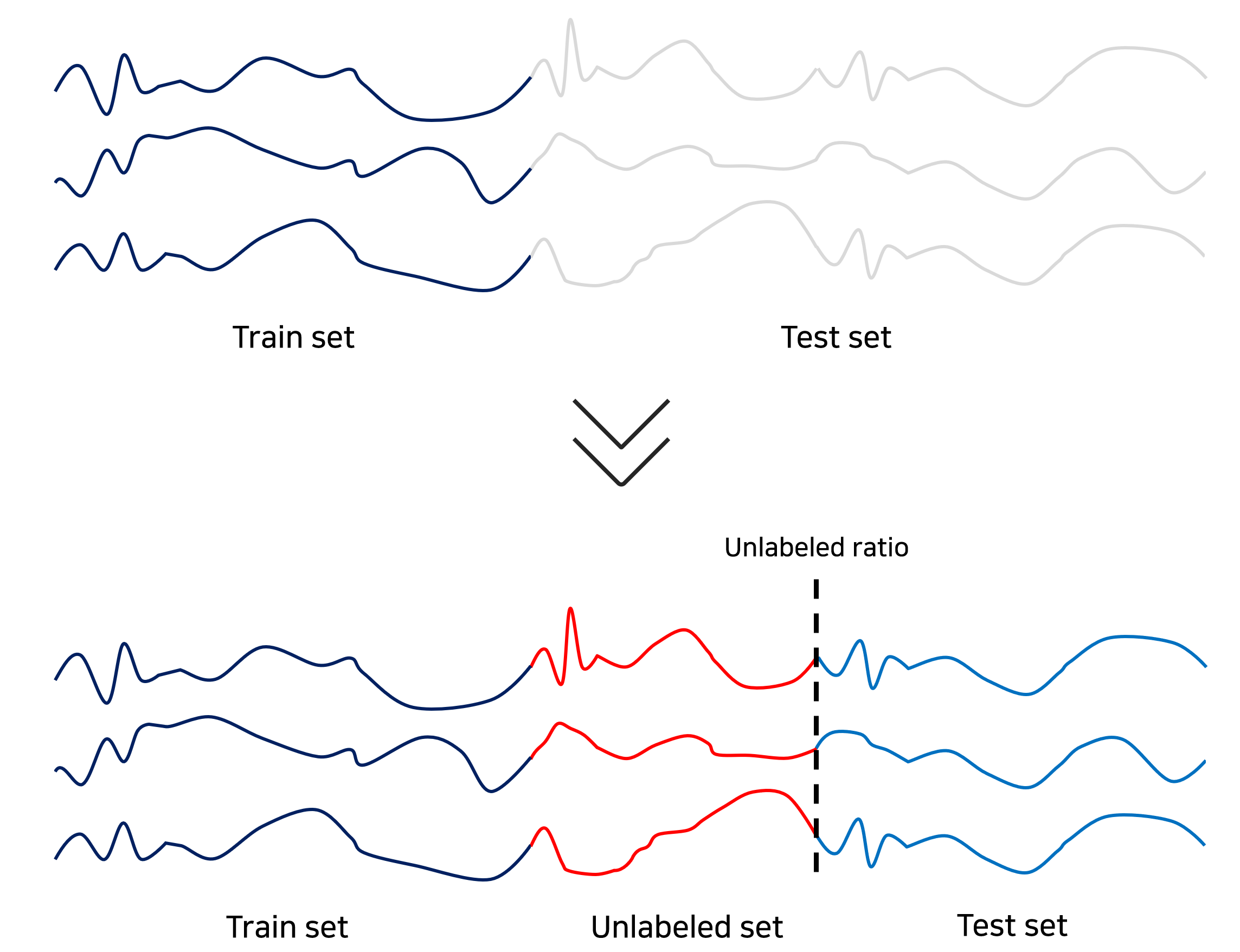}
    \caption{Example of dataset configuration process}
    \label{fig:dataset}
\end{figure*}

\textbf{Stage 2: Query Sampling.}
An unlabeled dataset, drawn from the test set, is used for active learning. This unlabeled set includes a mix of normal and anomalous sequences, but without labels. A query sampling strategy selects the most informative samples for annotation. In line with industrial environments, where large portions of data remain unlabeled \cite{AL_survey}, this stage mimics realistic conditions. The remainder of the test set, which includes ground-truth labels, is reserved for evaluation. The dataset configuration is summarized in \autoref{fig:dataset}.

\textbf{Stage 3: Supervised Feedback via Masked Reconstruction.}
The selected queries and their labels are used to refine the model through a masked time series reconstruction strategy. Specifically, random segments within each query sequence are masked, and the model is tasked with reconstructing these missing parts, a pretext task that encourages the learning of deeper temporal dependencies. To guide learning more effectively, a minimax objective is applied: the model minimizes reconstruction loss for normal queries while maximizing it for anomalous ones, enhancing its ability to discriminate between the two.

Stages 2 and 3 are repeated iteratively until a predefined query budget is reached, at which point the active learning process terminates.

\subsection{Training time series Anomaly Detection Model}
\label{sec:Training time series Anomaly Detection Model}
The first stage of the proposed framework involves training the backbone model in an unsupervised manner using only normal time series data. The model takes as input sequences $\mathbb{X} \in \mathbb{R}^{L \times C}$ and learns to reconstruct them with minimal error, producing outputs $\hat{\mathbb{X}}$ of the same dimensions. The training objective is to minimize the reconstruction loss, encouraging the model to capture the underlying temporal patterns characteristic of normal behavior.

This stage assumes that anomalous patterns, which deviate from the learned normal structure, will lead to noticeably higher reconstruction errors during inference. By exclusively training on normal data, the model is equipped to differentiate between normal and abnormal sequences based on reconstruction performance.

\subsection{Query Sampling Strategy}
\label{sec:Query Sampling Strategy}
To enhance the anomaly discrimination capability of the backbone model, we introduce a query sampling strategy that selects informative time series samples from an unlabeled dataset for additional supervised training. The pretrained model first computes a mean anomaly score for each sliding window in the unlabeled set, denoted as $\mathrm{score}_\mathrm{unlabeled}$, using the model's intrinsic scoring mechanism.

\begin{figure*}[!t]
    \centering  
    \includegraphics[width=14cm]{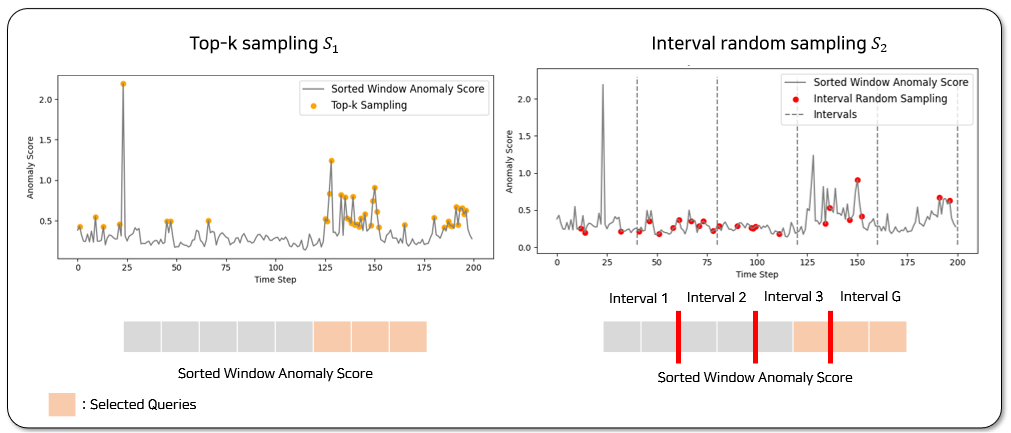}
    \caption{Illustration of two different ways of query sampling strategies}
    \label{fig:query_sampling}
\end{figure*}

As illustrated in \autoref{fig:query_sampling}, queries are selected in each iteration using a hybrid strategy $S$ that combines two sampling techniques—Top-$k$ Sampling $S_1$ and Interval Random Sampling $S_2$—following the methodology proposed in Active-MTSAD \cite{Active_MTSAD}.

\textbf{Top-$k$ Sampling ($S_1$)}: This method selects the top-$k$ windows with the highest anomaly scores from $\mathrm{score}_\mathrm{unlabeled}$, sorted in descending order. As shown in the left panel of \autoref{fig:query_sampling}, this sampling technique captures two critical cases: (1) noisy normal sequences that receive high anomaly scores, leading to false positives, and (2) genuine anomalies that are correctly ranked highly. Incorporating both types into supervised training allows the model to better distinguish between noisy normal behavior and true anomalies.

\begin{algorithm}
\caption{Query Sampling Strategy}
\label{alg:query sampling}
\begin{small}
\begin{algorithmic}[1]
\STATE \textbf{Input:} Trained backbone model $L$, Unlabeled dataset $U$
\STATE \textbf{Output:} Query set $Q$

\STATE Initialize $Q = \{\}$

\STATE score $\leftarrow$ Compute anomaly scores for all samples in $U$ with $L$ using \autoref{eq:anomaly_score}
\STATE $Sort(\text{score})$ \hfill \COMMENT{Descending order}

\STATE Select top k samples with the largest score to form set $S_1$  \hfill \COMMENT{top-k sampling}

\STATE Divide score into $G$ bins \hfill \COMMENT{Interval Random Sampling}
\FOR{each bin $b = 1$ to $G$}
\STATE $S_b \leftarrow \text{Randomly select $g$ samples from bin b} $ \\ 
\STATE $S_2 \leftarrow S_2 \cup S_b$ \\
\ENDFOR

\STATE $Q \leftarrow S_1 \cup S_2$
    
\STATE \textbf{Return:} Selected samples $Q$
\end{algorithmic}
\end{small}
\end{algorithm}

\textbf{Interval Random Sampling ($S_2$)}: To capture a broader range of patterns, this method first partitions the sorted anomaly scores into $G$ intervals and randomly samples $g$ windows from each, as illustrated in the right panel of \autoref{fig:query_sampling}. This approach ensures the inclusion of diverse samples, such as subtle anomalies with low anomaly scores that might otherwise be misclassified as normal. By doing so, $S_2$ complements $S_1$ by reducing false negatives and enhancing the model's robustness to variation in anomaly intensity.

To achieve comprehensive coverage of potential anomaly types, the final query set $Q$ is defined as the union of both strategies: $Q = S_1 \cup S_2$. In real-world industrial settings, the selected queries would require annotation by a human oracle. However, for experimental purposes, we assume that ground-truth labels are available immediately upon query selection. The full sampling procedure is detailed in Algorithm~\ref{alg:query sampling}.

\subsection{Masked time series Reconstruction based Feedback Strategy}
\label{sec:Masked time series Reconstruction based Feedback Strategy}

\begin{figure*}[!t]
    \centering  
    \includegraphics[width=14cm]{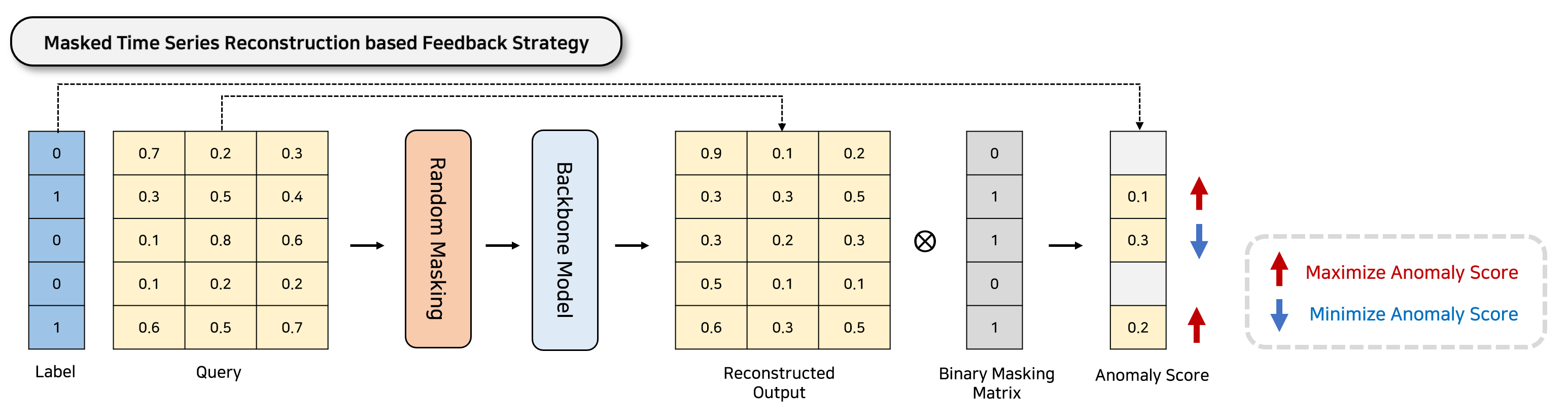}
    \caption{Illustration of Masked time series Reconstruction based Feedback Strategy}
    \label{fig:Feedback_strategy}
\end{figure*}

In the final stage of training, the backbone model is refined using a feedback strategy based on masked time series reconstruction and oracle-labeled queries. As illustrated in \autoref{fig:Feedback_strategy}, each selected multivariate query sequence is partially masked and reconstructed through the model. Supervised feedback is then applied via a minimax learning strategy, which treats normal and anomalous queries differently depending on their labels. Unlike Active-MTSAD \cite{Active_MTSAD}, which relies on pseudo-labels and metric learning, our method exclusively uses oracle-verified labels. This design avoids the propagation of labeling errors and improves robustness, especially in early active learning rounds where anomalies may be sparse or absent.

The feedback mechanism adopts the same sliding window configuration described in Section~\ref{sec:Query Sampling Strategy}. A fixed ratio of timesteps in each window is randomly masked, forming a pretext task that encourages the model to capture the temporal dynamics of the data, a technique validated in recent time series representation learning studies \cite{ts2vec, tst}.

The supervised learning objective adopts a minimax strategy based on the oracle-provided labels of query samples to make the discrepancy between normal and abnormal patterns more distinguishable. For each query at timestep $t$, a binary label $Y_t \in \{0, 1\}$ indicates whether the sample is normal ($Y_t = 0$) or anomalous ($Y_t = 1$). The model is trained to minimize the reconstruction error on masked timesteps for normal queries and to maximize it for anomalous queries.

This dual-objective strategy is defined as follows:
\begin{align} 
    \label{eq:Minimize_phase} 
    \text{Minimize Phase:} \quad \mathcal{L}_{\text{normal}} =\sum_{t=1}^{T}I_{\text{mask}}(t) \cdot I_{\text{normal}}(t) \cdot L(X_t, \hat{X}_t; \theta),
\end{align}
\begin{align} 
    \label{eq:Maximize_phase} 
    \text{Maximize Phase:} \quad \mathcal{L}_{\text{anomaly}} = -\sum_{t=1}^{T} I_{\text{mask}}(t) \cdot I_{\text{abnormal}}(t) \cdot L(X_t, \hat{X}_t; \theta).
\end{align}

Here, $T$ is the number of timesteps in the query window (i.e., the window size). $X_t$ and $\hat{X}_t$ denote the original and reconstructed values at time $t$, respectively. $\theta$ represents the model parameters, $L(\cdot)$ refers to the reconstruction loss, computed as the Mean Squared Error (MSE). $I_{\text{mask}}(t)$ is an indicator function (1 if timestep $t$ is masked, 0 otherwise). $I_{\text{normal}}(t) = \mathbb{I}(Y_t = 0)$ and $I_{\text{abnormal}}(t) = \mathbb{I}(Y_t = 1)$ are indicator functions selecting normal and anomalous samples, respectively.

By explicitly training the model to increase or decrease reconstruction error depending on the label, this minimax approach enhances its ability to differentiate between normal and anomalous sequences. Each training iteration integrates both phases and thus requires two backpropagation steps. Although a weighted combination of the two objectives was initially explored, the disjoint application of minimization and maximization yielded superior detection performance in empirical evaluation.

\subsection{Anomaly Detection}
\label{sec:Anomaly Detection}
Following model training, anomaly detection is performed by computing an anomaly score $S_t$ for each timestep $t$, based on the discrepancy between the input and reconstructed signals. This score reflects the likelihood that a given time point is anomalous. The anomaly score is defined as the mean reconstruction error across all channels, measured using the Euclidean distance:
\begin{align} 
    \label{eq:anomaly_score} 
    S_t = \frac{1}{C} \sum_{m=1}^{C} \sqrt{(X_{t,m} - \hat{X}_{t,m})^2}
\end{align}
Here, $X_{t,m}$ and $\hat{X}_{t,m}$ denote the original and reconstructed values, respectively, for channel $m$ at timestep $t$, and $C$ is the total number of channels. A sample at time $t$ is classified as anomalous if $S_t$ exceeds a predefined threshold. The threshold is selected based on the Best-F1 criterion, optimizing the balance between precision and recall on the validation or test set.

\section{Experimental Settings}
\label{sec:Experimental Settings}
\subsection{Datasets}
\label{Datasets}
To effectively evaluate the proposed active learning framework, it is essential to use datasets that contain diverse and independent anomalous segments, enabling the model to capture a broad range of anomaly patterns through iterative sampling. Among the five available time series anomaly detection benchmarks, we selected four datasets—SWaT \cite{SWaT}, PSM \cite{PSM}, Gecco \cite{NIPSdata,Gecco}, and Swan \cite{NIPSdata, Swan}, which provide sufficient temporal and structural variety in their anomaly instances. The univariate benchmark used in Section~\ref{sec:Motivation} was excluded from the main experiments due to its limited anomaly diversity, containing only a single anomaly segment.

\autoref{tab:dataset} summarizes the key characteristics of the selected datasets. Each dataset is partitioned as follows. The training set consists exclusively of normal time series data to support unsupervised pretraining. The unlabeled set is constructed by extracting the initial portion of the original test set based on a predefined unlabeled ratio. The remainder of the test set, containing both normal and anomalous samples with ground-truth labels, is used for evaluation. This data partitioning procedure is visually illustrated in \autoref{fig:dataset}.


\begin{table}[t!]
\centering
\caption{Details of time series anomaly detection benchmark datasets}
\vspace{0.1cm}
\label{tab:dataset}
\small 
\renewcommand{\arraystretch}{1.3} 
\setlength{\tabcolsep}{9pt} 
\begin{tabular}{c|ccccc}
\toprule
\textbf{Characteristics} & \textbf{SWaT} & \textbf{PSM} & \textbf{Gecco} & \textbf{Swan} & \textbf{UCR-001} \\
\midrule
\# of Variables & 51 & 25 & 9 & 38 & 1 \\
\# of Train & 396,000 & 132,481 & 69,260 & 60,000 & 14,113,128 \\
\# of Test (Labeled) & 449,919 & 87,841 & 69,261 & 60,000 & 18,109,080 \\
Anomaly (\%) & 12.0 & 27.8 & 1.1 & 32.6 & 1.4 \\
\bottomrule
\end{tabular}
\end{table}


\subsection{Baselines}
\label{sec:Baselines}
To evaluate the generalizability of the proposed framework and its ability to improve existing models, we adopt seven widely used reconstruction-based time series anomaly detection methods as backbone models: LSTM-VAE (2018) \cite{LSTM_VAE}, USAD (2020) \cite{USAD}, OmniAnomaly (2019) \cite{OmniAnomaly}, Transformer (2017) \cite{Transformer}, Anomaly Transformer (AT) (2022) \cite{anomaly_transformer}, VTT-SAT and VTT-PAT (2024) \cite{VTT}. In the case of Transformer, a multilayer perceptron (MLP) head is appended to the final encoder layer to align input and output dimensions. All models are trained by minimizing the reconstruction error between the input and the reconstructed output.

We also consider Active-MTSAD \cite{Active_MTSAD} as a relevant baseline due to its focus on active learning for anomaly detection under a similar problem setting. However, a direct comparison is not feasible for two key reasons. First, the official implementation is not publicly available, limiting reproducibility. Second, based on our re-implementation efforts, the framework halts in rounds where no anomalous queries are selected, an issue frequently encountered in realistic industrial datasets with sparse anomalies. This makes it incompatible with our evaluation protocol, which emphasizes continuous learning even in anomaly-scarce rounds. While Active-MTSAD provides valuable conceptual groundwork, these limitations underscore the practical advantage of our framework, which remains effective regardless of the anomaly distribution in the query rounds.

\subsection{Implementation Details}
\label{sec:Implementation Details}
We conduct a grid search over key hyperparameters to identify the configuration that yields the best F1 score. For Top-$k$ sampling, we test $k \in \{1, 5, 10, 15, 20\}$, and for Interval Random Sampling, the number of samples per bin is selected from $g \in \{1, 3, 5, 10\}$. The query budget is explored over $\{10, 50, 100\}$, and the unlabeled ratio is varied across $\{0.1, 0.2, 0.3, 0.4\}$. Additionally, the number of epochs used during feedback training—referred to as feedback epochs—is selected from $\{10, 50, 100\}$.

All experiments use a non-overlapping sliding window of fixed size 50 across all datasets. We adopt the Adam optimizer \cite{adam} with a learning rate of $10^{-2}$, and use a batch size of 8. Early stopping is applied with a patience of 10 epochs. Hyperparameters specific to each backbone model are configured according to the original implementations in their respective papers.

The active learning process proceeds iteratively until the cumulative number of annotated queries reaches the predefined budget. All experiments are implemented in PyTorch \cite{pytorch} and executed on an NVIDIA GeForce RTX 2080 Ti BLOWER 11GB GPU.

\subsection{Evaluation Metrics}
\label{sec:Evaluation Metrics}
We evaluate model performance using three widely adopted metrics for time series anomaly detection: Point-wise F1 score (F1), Point-Adjusted F1 score ($\text{F1}_\text{PA}$), and AUC (Area Under the PA\%K Curve).

The Point-Adjusted F1 score ($\text{F1}_\text{PA}$) considers an entire anomaly segment to be correctly detected if at least one point within the segment is identified as anomalous \cite{anomaly_transformer, VTT}. This metric reflects a practical assumption in industrial settings, where a single fault often affects multiple consecutive timestamps. However, it has been shown to overestimate detection performance in certain cases \cite{PAK}.

To address this limitation, the PA\%K metric was introduced as a refined evaluation method \cite{PAK}. It applies the PA adjustment only if the detected anomalies cover at least K\% of the ground-truth anomaly segment. In this study, we compute the AUC by varying K from 0 to 100 (in increments of 1) and measuring the area under the resulting PA\%K curve. This AUC provides a more comprehensive and robust evaluation of anomaly detection performance across varying levels of detection coverage.

\subsection{Best-F1 Threshold Searching}
\label{sec:Best-f1}
A significant challenge in evaluating unsupervised time series anomaly detection models lies in establishing a classification threshold. Since these models inherently produce anomaly scores rather than discrete labels, the choice of a specific thresholding technique can itself become a confounding variable, influencing the final performance metrics.
In this study, our primary objective is to assess the intrinsic anomaly detection capacity of each model that is, the quality of the anomaly scores they generate. Therefore, to ensure a fair and direct comparison, we adhere to a standard benchmarking protocol used in time series anomaly detection literature. For our proposed model and all baseline models, we evaluate the anomaly scores generated for the test dataset against all possible threshold values. The performance is then reported based on the threshold that yields the maximum F1-score, i.e., Best-F1. This approach ensures that all models are compared under their most optimal conditions, allowing their core detection capabilities to be evaluated fairly without performance being skewed by the choice of a sub-optimal thresholding heuristic \cite{VTT}.

\section{Experimental Results}
\label{sec:Experimental Results}

\subsection{Main Results}
\label{sec:Main Results}
To verify that the performance gains of the proposed framework are not simply attributable to increased exposure to normal data, we conduct comparative experiments under different training set configurations. Specifically, we compare three setups: (1) the backbone model trained only on the original training set (\textit{w/o unlabeled}), (2) trained with both the training set and the unlabeled set (\textit{w/ unlabeled}), and (3) trained using the proposed active learning framework (\textit{Ours}). In the \textit{Ours} configuration, the backbone model is first trained in an unsupervised manner (\textit{w/o unlabeled}) and then fine-tuned using our active learning framework. 

\begin{figure*}[t!]
    \centering  
    \includegraphics[width=14cm]{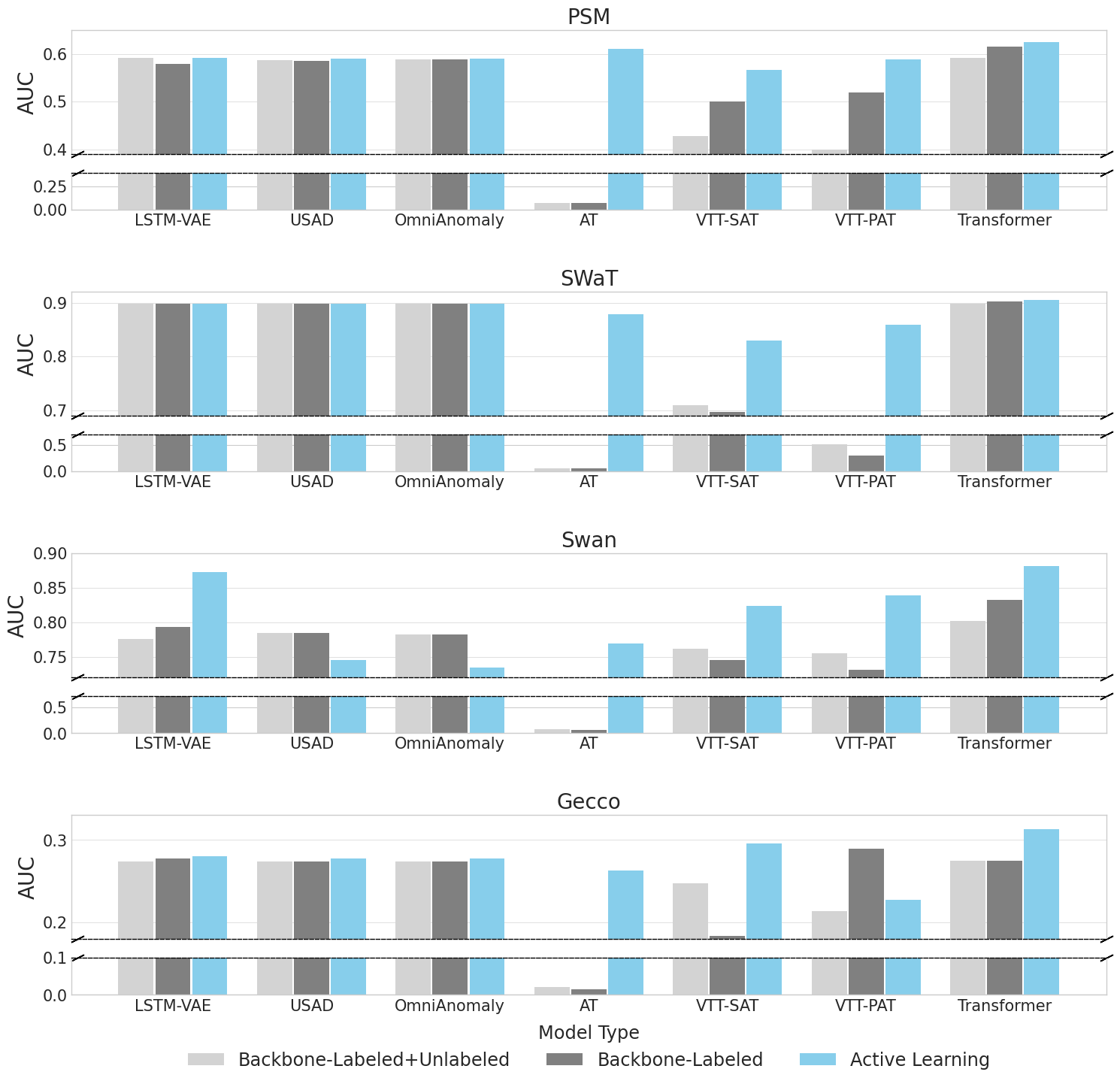}
    \caption{Bar plot showing AUC of experimental result with seven different backbone models on each benchmark. Backbone-Labeled+Unlabeled corresponds to ${w/}$ ${unlabeled}$, Backbone-Labeled corresponds to ${w/o}$ ${unlabeled}$ and Active Learning corresponds to $Ours$ in \autoref{tab:main_exp}}
    \label{fig:barplot}
\end{figure*}

Detailed results are presented in \autoref{tab:main_exp}. Bolded values in the table indicate the best performance across all backbone models for each dataset and metrics. To clarify the gains, \textbf{average diff.} in \autoref{tab:average_diff} quantifies the average improvement in AUC between the \textit{w/ unlabeled} model and the same model enhanced by active learning, across the four benchmark datasets. The consistently higher bars (light blue) in \autoref{fig:barplot} for the active learning models highlight the effectiveness of the proposed framework. Notably, the Transformer with active learning achieves the greatest overall improvement, with an average AUC increase of 3.95\%.
Moreover, this configuration achieves the top rank in 11 out of 12 evaluation scenarios (4 datasets $\times$ 3 metrics), and secures 10 first-place results out of 21 model variations, covering seven backbones across three training setups. This suggests that attention-based architectures, such as the Transformer, are especially well-suited for feedback-based learning due to their strength in capturing temporal dependencies.

\begin{table}[t!]
\caption{Experiment results on each benchmark for seven different backbone models}
\vspace{0.1cm}
\setlength{\tabcolsep}{6pt} 
\renewcommand{\arraystretch}{1.4} 
\centering
\Huge
\resizebox{\columnwidth}{!}{%
{
\begin{tabular}{c|c|ccc|ccc|ccc|ccc}
\Xhline{2.5pt}
\multicolumn{2}{c|}{\Huge{\textbf{Dataset}}} & 
  \multicolumn{3}{c|}{\Huge{\textbf{PSM}}} &
  \multicolumn{3}{c|}{\Huge{\textbf{SWaT}}} &
  \multicolumn{3}{c|}{\Huge{\textbf{Gecco}}} &
  \multicolumn{3}{c}{\Huge{\textbf{Swan}}} \\  
\hline
\multicolumn{1}{c|}{\Huge{\textbf{Backbone}}} &
  {\Huge{\textbf{Type}}} &
  \Huge{\textbf{$\text{F1}_\text{PA}$}} &
  \Huge{\textbf{F1}} &
  \Huge{\textbf{AUC}} &
  \Huge{\textbf{$\text{F1}_\text{PA}$}} &
  \Huge{\textbf{F1}} &
  \Huge{\textbf{AUC}} &
  \Huge{\textbf{$\text{F1}_\text{PA}$}} &
  \Huge{\textbf{F1}} &
  \Huge{\textbf{AUC}} &
  \Huge{\textbf{$\text{F1}_\text{PA}$}} &
  \Huge{\textbf{F1}} &
  \Huge{\textbf{AUC}} \\
\cmidrule{1-14}
\morecmidrules
\cmidrule{1-14}

\multicolumn{1}{c|}{} &
  \textit{w/ unlabeled} &
  0.9311 &
  0.5394 &
  0.5914 &
  0.9210 &
  0.8463 &
  0.8976 &
  0.3002 &
  0.2168 &
  0.2739 &
  0.8915 &
  0.6132 &
  0.7763 \\
\multicolumn{1}{c|}{\textbf{LSTM-VAE}\cite{LSTM_VAE}} &
  \textit{\textit{w/o unlabeled}} &
  \textbf{0.9332} &
  0.5259 &
  0.5789 &
  0.9215 &
  0.8463 &
  0.8976 &
  0.3008 &
  0.2209 &
  0.2772 &
  0.8918 &
  0.6494 &
  0.7938 \\
\multicolumn{1}{c|}{} &
  \textit{Ours} &
  0.9314 &
  \textbf{0.5401} &
  \textbf{0.5917} &
  \textbf{0.9235} &
  \textbf{0.8463} &
  \textbf{0.8978} &
  \textbf{0.3014} &
  \textbf{0.2249} &
  \textbf{0.2801} &
  \textbf{0.8988} &
  \textbf{0.7914} &
  \textbf{0.8736} \\ \hline
\multicolumn{1}{c|}{} &
  \textit{w/ unlabeled} &
  0.8734 &
  \textbf{0.5293} &
  0.5878 &
  \textbf{0.9231} &
  0.8463 &
  \textbf{0.8976} &
  0.3002 &
  0.2168 &
  0.2739 &
  \textbf{0.8914} &
  \textbf{0.6279} &
  0.7847 \\
\multicolumn{1}{c|}{\textbf{USAD}\cite{USAD}} &
  \textit{w/o unlabeled} &
  0.8857 &
  0.5270 &
  0.5861 &
  0.9132 &
  0.8463 &
  0.8976 &
  0.3002 &
  0.2168 &
  0.2744 &
  \textbf{0.8914} &
  \textbf{0.6279} &
  \textbf{0.7847} \\
\multicolumn{1}{c|}{} &
  \textit{Ours} &
  \textbf{0.9297} &
  0.5224 &
  \textbf{0.5896} &
  0.9113 &
  \textbf{0.8463} &
  0.8975 &
  \textbf{0.3008} &
  \textbf{0.2209} &
  \textbf{0.2777} &
  0.8902 &
  0.6064 &
  0.7453 \\ \hline
\multicolumn{1}{c|}{} &
  \textit{w/ unlabeled} &
  \textbf{0.9300} &
  \textbf{0.5369} &
  0.5888 &
  \textbf{0.9224} &
  0.8463 &
  \textbf{0.8976} &
  0.3002 &
  0.2168 &
  0.2739 &
  \textbf{0.8916} &
  \textbf{0.6220} &
  \textbf{0.7821} \\
\multicolumn{1}{c|}{\textbf{OmniAnomaly}\cite{OmniAnomaly}} &
  \textit{w/o unlabeled} &
  0.9300 &
  0.5369 &
  0.5888 &
  0.9223 &
  0.8463 &
  0.8976 &
  0.3002 &
  0.2168 &
  0.2739 &
  \textbf{0.8916} &
  \textbf{0.6220} &
  0.7821 \\
\multicolumn{1}{c|}{} &
  \textit{Ours} &
  0.9220 &
  0.5369 &
  \textbf{0.5907} &
  0.9222 &
  \textbf{0.8463} &
  0.8976 &
  \textbf{0.3008} &
  \textbf{0.2209} &
  \textbf{0.2777} &
  0.8905 &
  0.5855 &
  0.7340 \\ \hline
\multicolumn{1}{c|}{} &
  \textit{w/ unlabeled} &
  0.9319 &
  0.5402 &
  0.5918 &
  \textbf{0.9206} &
  0.8463 &
  0.8976 &
  0.3002 &
  0.2168 &
  0.2749 &
  0.8898 &
  0.6629 &
  0.8026 \\
\multicolumn{1}{c|}{\textbf{Transformer}\cite{Transformer}} &
  \textit{w/o unlabeled} &
  0.8897 &
  0.5641 &
  0.6152 &
  0.9181 &
  0.8514 &
  0.9023 &
  0.3002 &
  0.2168 &
  0.2749 &
  0.8902 &
  0.7258 &
  0.8330 \\
\multicolumn{1}{c|}{} &
  \textit{Ours} &
  \textbf{0.9324} &
  \textbf{0.5653} &
  \textbf{0.6248} &
  0.9108 &
  \textbf{0.8530} &
  \textbf{0.9044} &
  \textbf{0.3136} &
  \textbf{0.3067} &
  \textbf{0.3135} &
  \textbf{0.9057} &
  \textbf{0.8010} &
  \textbf{0.8821} \\ \hline
\multicolumn{1}{c|}{} &
  \textit{w/ unlabeled} &
  0.9720 &
  0.0500 &
  0.0759 &
  \textbf{0.9728} &
  0.0426 &
  0.0672 &
  0.2795 &
  0.0126 &
  0.0219 &
  0.8381 &
  0.0500 &
  0.0725 \\
\multicolumn{1}{c|}{\textbf{AT}\cite{anomaly_transformer}} &
  \textit{w/o unlabeled} &
  \textbf{0.9725} &
  0.0470 &
  0.0717 &
  0.9608 &
  0.0374 &
  0.0611 &
  0.2730 &
  0.0082 &
  0.0148 &
  0.8442 &
  0.0477 &
  0.0706 \\
\multicolumn{1}{c|}{} &
  \textit{Ours} &
  0.9240 &
  \textbf{0.5246} &
  \textbf{0.6113} &
  0.9239 &
  \textbf{0.7847} &
  \textbf{0.8783} &
  \textbf{0.3037} &
  \textbf{0.1971} &
  \textbf{0.2627} &
  \textbf{0.8798} &
  \textbf{0.6150} &
  \textbf{0.7696} \\ \hline
\multicolumn{1}{c|}{} &
  \textit{w/ unlabeled} &
  0.8606 &
  0.4044 &
  0.4285 &
  \textbf{0.9280} &
  0.4815 &
  0.7097 &
  0.2911 &
  0.1513 &
  0.2473 &
  0.8433 &
  0.6084 &
  0.7621 \\
\multicolumn{1}{c|}{\textbf{VTT-SAT}\cite{VTT}} &
  \textit{w/o unlabeled} &
  0.8606 &
  0.4044 &
  0.5000 &
  \textbf{0.9280} &
  0.4815 &
  0.6977 &
  0.2868 &
  0.1186 &
  0.1838 &
  0.8434 &
  0.6094 &
  0.7451 \\
\multicolumn{1}{c|}{} &
  \textit{Ours} &
  \textbf{0.9332} &
  \textbf{0.4146} &
  \textbf{0.5661} &
  0.9218 &
  \textbf{0.7079} &
  \textbf{0.8300} &
  \textbf{0.3056} &
  \textbf{0.2536} &
  \textbf{0.2957} &
  \textbf{0.8466} &
  \textbf{0.7197} &
  \textbf{0.8241} \\ \hline
\multicolumn{1}{c|}{} &
  \textit{w/ unlabeled} &
  \textbf{0.9392} &
  0.4076 &
  0.4000 &
  0.8683 &
  0.3319 &
  0.5129 &
  0.2900 &
  0.1431 &
  0.2142 &
  0.8434 &
  0.6080 &
  0.7551 \\
\multicolumn{1}{c|}{\textbf{VTT-PAT} \cite{VTT}} &
  \textit{w/o unlabeled} &
  \textbf{0.9392} &
  0.4076 &
  0.5201 &
  \textbf{0.9471} &
  0.2823 &
  0.2972 &
  \textbf{0.3037} &
  \textbf{0.2413} &
  \textbf{0.2894} &
  0.8434 &
  0.6080 &
  0.7315 \\
\multicolumn{1}{c|}{} &
  \textit{Ours} &
  0.9312 &
  \textbf{0.5375} &
  \textbf{0.5894} &
  0.9268 &
  \textbf{0.7269} &
  \textbf{0.8586} &
  0.2922 &
  0.1595 &
  0.2276 &
  \textbf{0.8478} &
  \textbf{0.7881} &
  \textbf{0.8394} \\ \Xhline{2.5pt}
\end{tabular}%
}
}
\centering
{\small \textbf{Bold}: Best result among each backbone and benchmark group}
\label{tab:main_exp}
\end{table}

\begin{table}[t!]
\centering
\caption{Average performance difference (\textit{Ours} -- Baseline) across datasets for each backbone}
\vspace{0.1cm}
\label{tab:average_diff}
\resizebox{\textwidth}{!}{%
\large  
\setlength{\tabcolsep}{10pt}
\renewcommand{\arraystretch}{1.3}
\begin{tabular}{l|ccccccc}
\Xhline{1pt}
\textbf{Metric} & LSTM-VAE & USAD & OmniAnomaly & Transformer & AT & VTT-SAT & VTT-PAT \\
\Xhline{0.8pt}
\textbf{$\text{F1}_\text{PA}$} & 0.0028 & 0.0110 & -0.0022 & 0.0050 & -0.0077 & 0.0210 & 0.0143 \\
\textbf{F1}     & 0.0468 & -0.0061 & -0.0081 & 0.0650 & 0.4916 & 0.1125 & 0.1804 \\
\textbf{AUC}    & 0.0260 & -0.0085 & -0.0106 & 0.0395 & 0.5711 & 0.0921 & 0.1582 \\
\Xhline{1pt}
\end{tabular}%
}
\end{table}

The Anomaly Transformer (AT) also benefits significantly from our framework, particularly in narrowing the gap between PA and point-wise F1 scores. It shows an average improvement of 49.16\% in F1 and 57.11\% in AUC. Likewise, VTT-SAT and VTT-PAT which modify the standard Transformer attention with temporal-variable attention—achieve AUC gains of 9.21\% and 15.82\%, respectively.

However, performance degradation is observed for USAD and OmniAnomaly in some datasets. These models incorporate adversarial or probabilistic learning objectives beyond reconstruction loss, which may conflict with the minimax-based feedback strategy. This indicates that the proposed method is more effective when aligned with models that rely primarily on reconstruction-based objectives.

While comparing the \textit{w/ unlabeled} and \textit{w/o unlabeled} training configurations, we observe that performance trends are not always consistent across models. For instance, USAD and OmniAnomaly sometimes perform better when trained with additional unlabeled data than when enhanced with active learning, whereas the opposite is true in other configurations. These inconsistencies suggest that simply increasing the training data volume does not guarantee improved performance. Instead, the observed gains from our method are attributable to the quality of the selected queries and the effectiveness of the feedback strategy, rather than the quantity of training data alone.

\begin{figure*}[!t]
    \centering  
    \includegraphics[width=14cm]{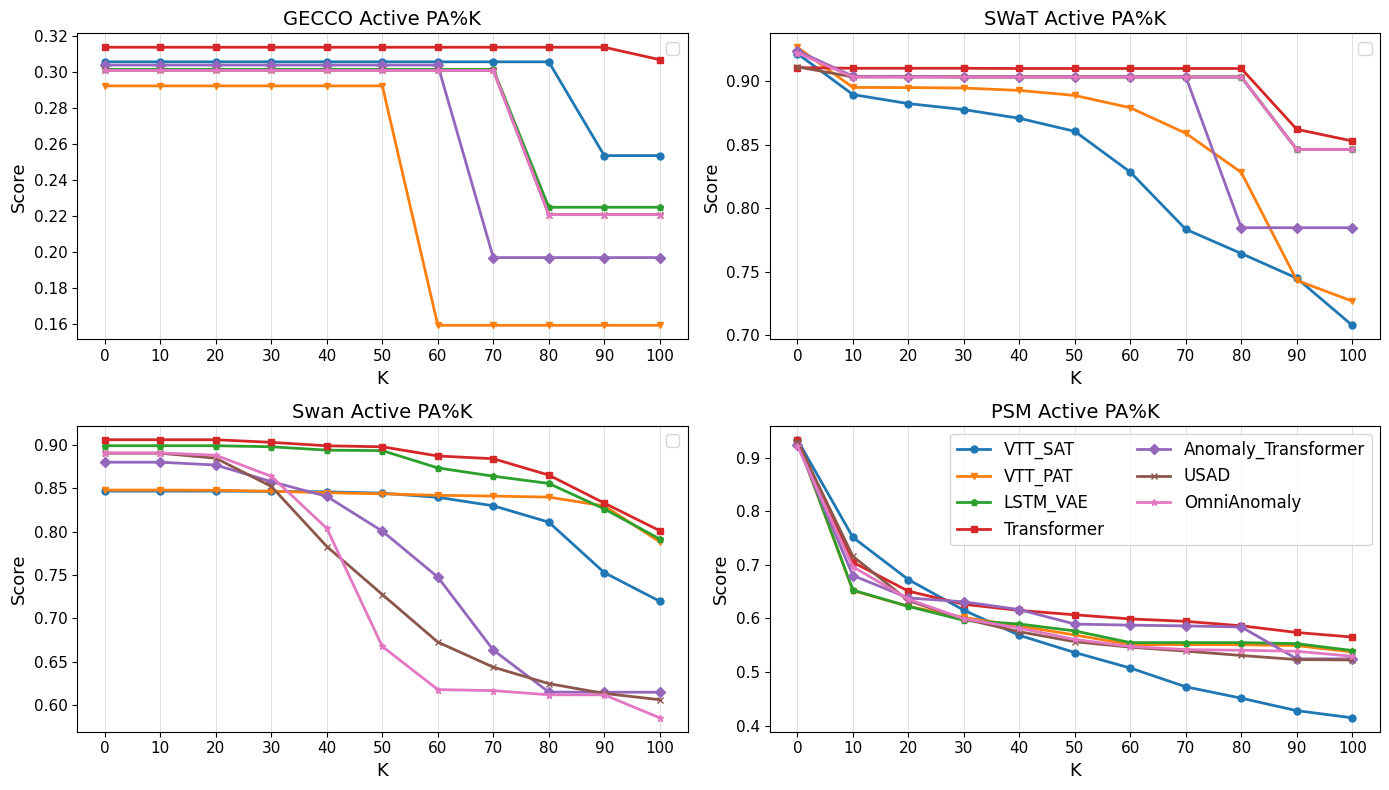}
    \caption{PA\%K curve for four datasets and seven backbone models with our proposed framework applied}
    \label{fig:AUC}
\end{figure*}

\autoref{fig:AUC} shows the PA\%K curves across different models and datasets. For most datasets, PA\%K values decline gradually as $K$ increases, except for Gecco, where shorter anomaly segments lead to steeper drops in performance. Despite this, models enhanced with active learning show stable detection performance across a wide range of $K$ values.

Lastly, the Anomaly Transformer demonstrates high sensitivity to $K$ due to its reliance on association discrepancy in anomaly scoring. While this makes it effective at detecting long-duration anomalies, it can reduce precision in point-wise evaluations.

A notable observation from Table 2 is that while our framework significantly improves Transformer-based models on e.g., Swan, Gecco and SWaT, the performance gain on PSM is marginal. This is not a flaw in the framework, but rather an assumption mismatch between our Active Learning strategy and the unique characteristics of the PSM dataset.
Our framework, like most active learning strategies for anomaly detection, is designed for scenarios where anomalies in the unlabeled set are rare and ambiguous as in SWaT, Swan, allowing the query strategy to efficiently concentrate these high-value `noisy normal' and `subtle anomaly' samples from a large pool of normal data.
However, as the PSM unlabeled dataset consists of 31\% anomalies, the fundamental premise of Active Learning is violated. This data saturation leads to a critical failure in the query strategy. First failure is a biased Feedback where query sampling strategy, which is not designed for such anomaly-abundant data, selects a small and non-representative biased subset from a massive, diverse anomaly class. Second failure is ab overfitting of Feedback. Our feedback strategy trains the model to maximize error on these specific anomalous queries. However, as these queries do not represent the full diversity of anomalies in PSM, the model may overfit to reject these specific patterns while failing to generalize to other anomaly types in the test set.

Therefore, the marginal gain on PSM is due to the dataset's violation of the ``rare anomaly" assumption, which in turn limits the generalization potential of our robust feedback mechanism.

\subsection{Ablation Study}
\label{sec:sec5.3}

\begin{figure*}[!t]
    \centering  
    \includegraphics[width=14cm]{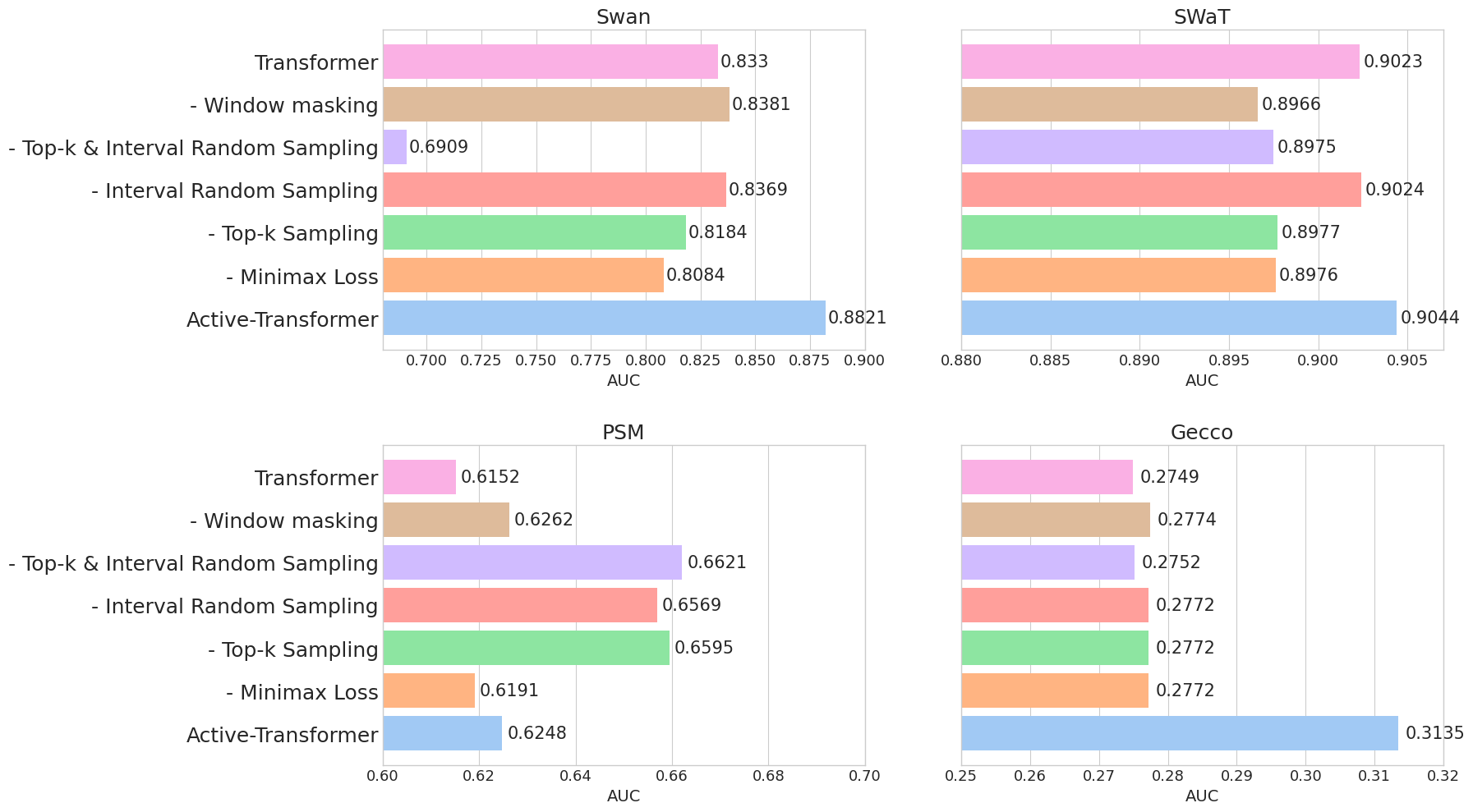}
    \caption{AUC score of Transformer backbone model for four datasets when major components of proposed framework are removed individually}
    \label{fig:ablation}
\end{figure*}

To evaluate the individual contribution of each component in the proposed framework, we conduct an ablation study using the Transformer, which showed the strongest performance in the main experiments. We define the following ablation settings:
\begin{itemize}
    \item \textit{-Minimax Loss}: The minimax loss in the feedback strategy is replaced with a weighted sum of reconstruction losses for normal and anomalous queries.
    \item \textit{-Top-k sampling}/\textit{-Interval Random Sampling}: Each sampling method ($S_1$ or $S_2$) is independently replaced with uniform random sampling of the same size.
    \item \textit{-Top-k \& Interval Random Sampling}: All queries are randomly sampled in each round, effectively removing the designed sampling strategy.
    \item \textit{-Window Masking}: The masking pretext task is removed; the model directly reconstructs raw sequences without any masked input.
\end{itemize}

\autoref{fig:ablation} presents the results of the ablation study across the benchmark datasets. Overall, applying all components of the proposed framework yields the highest performance across all datasets except for the PSM. The analysis reveals the following key findings:
\begin{itemize}
    \item Query Sampling Strategy: Replacing either $S_1$ (top-$k$) or $S_2$ (interval sampling) with random sampling consistently leads to performance degradation or marginal gains over the backbone model. This confirms that our proposed feedback mechanism is highly synergistic with a targeted query strategy. 
    \item Masking Pretext Task: Removing the masking task from the feedback strategy results in a notable decline in performance. This underscores the importance of learning temporal dependencies through masked reconstruction, aligning with findings from time series representation learning literature.
    \item Minimax Loss: Substituting the minimax objective with a simple weighted sum leads to a significant drop in AUC. This result strongly validates our hypothesis, aligning with findings from \cite{anomaly_transformer}, that simply reconstructing samples is insufficient. The minimax objective's role in actively maximizing the error for anomalies while minimizing it for normals is crucial for making the reconstruction discrepancy distinguishable. The performance drop confirms that treating normal and anomalous queries with distinct, opposing optimization goals is essential for robust discrimination.
\end{itemize} 

These results collectively highlight the effectiveness and necessity of each component in the proposed framework. The synergy between targeted sampling, masking-based feedback, and asymmetric loss formulation is crucial to achieving reliable performance improvements in active learning-based time series anomaly detection.

\section{Analysis}
\label{sec:Analysis}

\subsection{Effect of Active Learning on Backbone model}
\label{sec:Effect of Active Learning on Backbone model}

\begin{figure*}[!t]
    \centering  
    \includegraphics[width=14cm]{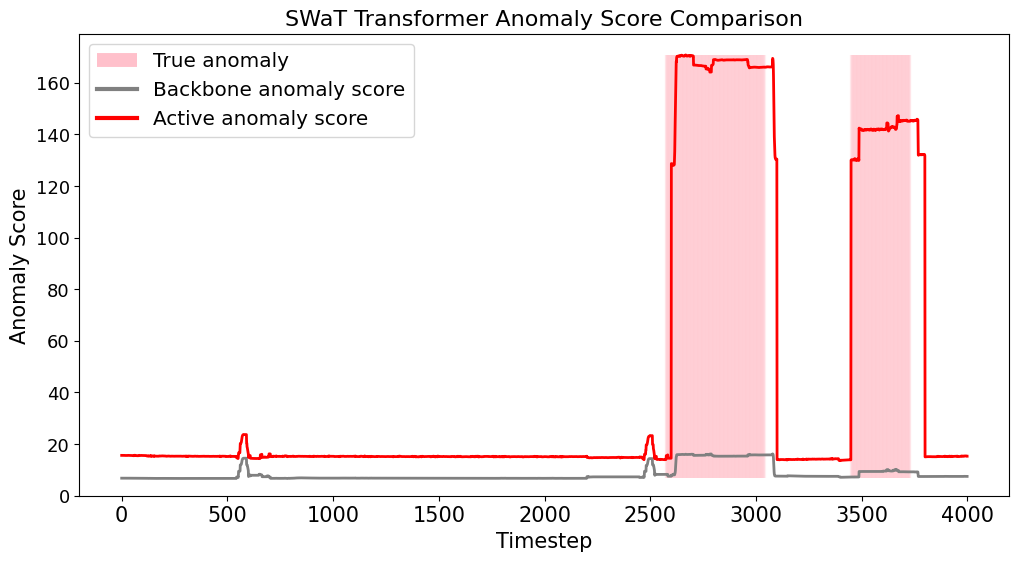}
    \caption{Comparison of anomaly score of Transformer backbone model when active learning is implemented on SWaT test set from timestep 255,000 to 259,000}
    \label{fig:active_case}
\end{figure*}

\autoref{fig:active_case} compares the anomaly scores generated by the original Transformer backbone and the active Transformer, which incorporates the proposed framework, for the SWaT dataset. As shown in the figure, the active Transformer produces significantly elevated anomaly scores within anomalous segments, improving the separation between normal and abnormal patterns. Although a slight increase in scores is also observed in normal segments, the contrast between the two regions becomes more distinct, reducing false negatives. This suggests that our framework enhances the model’s ability to detect subtle anomalies by refining its sensitivity to abnormal patterns through supervised feedback.

\subsection{Budget Size for Active Learning}
\label{sec:Budget Size for Active Learning}

\begin{table}[!t]
\centering
\caption{Size of each four benchmark and portion that each corresponding budget size takes from unlabeled set of each benchmark}
\label{tab:budget_size}
\vspace{0.2cm} 
\small 
\setlength{\tabcolsep}{4pt} 
\renewcommand{\arraystretch}{1.3} 
\begin{tabular}{c|ccc|c|c|c}
\toprule
\multirow{2}{*}{\textbf{Dataset}} &
  \multicolumn{3}{c|}{\textbf{Budget Size (windows)}} &
  \multirow{2}{*}{\begin{tabular}[c]{@{}c@{}}\# of Train\\ Timesteps\end{tabular}} &
  \multirow{2}{*}{\begin{tabular}[c]{@{}c@{}}\# of Test\\ Timesteps\end{tabular}} &
  \multirow{2}{*}{\begin{tabular}[c]{@{}c@{}} \# of windows \\ in unlabeled set\end{tabular}} \\ 
\cmidrule(lr){2-4}
      & \multicolumn{1}{c}{30} & \multicolumn{1}{c}{50} & \multicolumn{1}{c|}{100} &         &         &       \\ 
\midrule
PSM   & 4.27\% & 7.12\%                 & {\ul \textbf{14.25\%}} & 132,481 & 87,841  & 702   \\
SWaT  & 0.83\% & {\ul \textbf{1.39\%}}  & 2.78\%                 & 495,000 & 449,919 & 3,599 \\
Gecco & 5.42\% & {\ul \textbf{9.03\%}}  & 18.05\%                & 69,260  & 69,261  & 554   \\
Swan  & 6.25\% & {\ul \textbf{10.42\%}} & 20.83\%                & 60,000  & 60,000  & 480   \\ 
\bottomrule
\end{tabular}
\end{table}

To assess the trade-off between annotation effort and detection performance, we evaluate three different budget sizes: 30, 50, and 100 queries. \autoref{tab:budget_size} reports the relative proportion of each budget compared to the total number of windows in the unlabeled set for each dataset. The optimal budget for the active Transformer, in terms of F1 score, is highlighted in bold. 

As the results show, annotating as little as 21\% of the unlabeled windows can yield substantial performance gains. For example, the Transformer backbone achieves the best results with a budget of 50 in SWaT, Gecco, and Swan, and 100 in PSM. These figures correspond to only 1\% to 14\% of the total unlabeled windows; demonstrating the framework’s ability to deliver strong improvements with minimal annotation cost.

\subsection{Statistics of Selected Queries}
\label{sec:Statistics of Selected Queries}

\begin{table}[!t]
\centering
\caption{Query sampling result with set of hyperparameters that yields the highest AUC for Transformer-backboned active learning}
\vspace{0.1cm} 
\small 
\setlength{\tabcolsep}{4pt} 
\renewcommand{\arraystretch}{1.2} 
\begin{tabular}{c|c|c|c|c}
\toprule
\textbf{Dataset} & \textbf{Anomaly ratio} & \textbf{Budget} & \textbf{\begin{tabular}[c]{@{}c@{}}Anomaly ratio \\ in unlabeled set\end{tabular}} & \textbf{\begin{tabular}[c]{@{}c@{}} \# of windows  \\ in unlabeled set\end{tabular}} \\ 
\midrule
PSM   & 9\%  & 30 & 31\% & 702  \\ 
SWaT  & 45\% & 50 & 6\%  & 3,599 \\ 
Gecco & 22\% & 50 & 2\%  & 554  \\ 
Swan  & 85\% & 50 & 16\% & 480  \\ 
\bottomrule
\end{tabular}
\label{tab:query-result}
\end{table}

\autoref{tab:query-result} shows the proportion of anomalies among the queries selected by the active learning strategy in each round when using the Transformer as the backbone. Notably, across three datasets, the fraction of anomalous queries exceeds the anomaly ratio in the entire unlabeled set. This outcome indicates that the query sampling strategy, combining top-$k$ selection and interval-based random sampling, successfully prioritizes informative samples within a limited budget. Moreover, the sampling maintains a balance by capturing both noisy normal data and genuine anomalies, which is essential for effective supervised feedback and model refinement.

\section{Conclusion}
\label{sec:Conclusion}

Time series anomaly detection plays a vital role in domains such as manufacturing, finance, and healthcare, where the timely identification of anomalies is crucial for preventing operational disruptions and financial losses. While unsupervised learning has been widely adopted to mitigate the scarcity of labeled data, existing methods often struggle to capture the complex temporal dependencies in multivariate sequences. In particular, they frequently misclassify noisy normal data as anomalies and fail to detect subtle anomalous patterns—limiting their practical effectiveness.

To address these challenges, we proposed an active learning-based framework that enhances reconstruction-based unsupervised anomaly detection models by selectively incorporating supervised feedback. The framework introduces a dual query sampling strategy, combining top-$k$ selection and interval-based random sampling, to obtain a diverse set of informative queries from unlabeled data. This sampling approach ensures the inclusion of both noisy normal sequences and near-normal anomalies.

To refine model discrimination, we introduced a masked reconstruction-based feedback mechanism guided by oracle-provided labels. Using a minimax loss, the model is trained to minimize reconstruction error for normal queries while maximizing it for anomalies, improving its ability to distinguish between the two. The framework is model-agnostic and demonstrated its effectiveness across seven backbones and four datasets, achieving AUC improvements in 82\% of the 28 evaluation cases. Notably, when applied to a Transformer encoder, the framework achieved an average AUC gain of 7.56 percentage points, surpassing all baseline methods.

While the proposed method shows strong performance, its reliance on the minimax objective may introduce learning instability under certain conditions. Future work will investigate alternative feedback strategies, such as those based on metric learning without pseudo-labels, to improve training stability and generalization. Moreover, given the framework’s sensitivity to hyperparameter settings, future research should aim to develop more robust and adaptive mechanisms to ensure consistent performance across datasets and model configurations.

\clearpage
\bibliographystyle{elsarticle-num-names}
\bibliography{AL_references}
\end{document}